# AI WITH ALIEN CONTENT AND ALIEN METASEMANTICS

Herman Cappelen (*The University of Hong Kong*) and Josh Dever (*The University of Texas at Austin*)

# 1. The Semantics and Metasemantics of AI: An Introduction

In earlier work (Cappelen and Dever 2021) we take metasemantic views that have been proposed for thinking about content in humans and try to apply them to AI systems. Along the way we encounter aspects of these anthropocentric metasemantic views that are ill-suited to AI systems. One response to that was to try to build more abstracted versions of those metasemantic features, so that some of the details that are highly specific to humans get taken out and replaced by tools that are more broadly applicable (that was the strategy in our 2021 book). In this paper we want to look at another response. Maybe AIs have alien contents and we need a fundamentally different kind of metasemantics for AI systems than we do for humans. The aim is to explore a cluster of closely related topics:

> (i) What would it be for a system to have alien (non-human) contents and how could we recognize these as contents?
> (ii) What would it be for a system to have alien (non-human) metasemantics and how could we recognize it as a metasemantics?
> (ii) Could artificial intelligences have alien contents and alien metasemantics?
> (iii) If AIs have alien contents and metasemantics, how do we communicate (and otherwise engage) with them?

Much of what we have to say is exploratory. This is in large part an effort to convince readers that it's possible that AIs have both alien contents and alien metasemantics. To the extent that we have conclusions they are the following:

> **Conclusion 1**. AIs could have alien contents and alien metasemantics.
> **Conclusion 2**. Even if they do, we can find ways to communicate with them.
> **Conclusion 3**. Taking these options seriously is needed to think clearly about issues such as existential risk, the value alignment problem, and explainable AI.

A larger goal in this paper (and also in Cappelen and Dever 2021) is to illustrate how familiar work from philosophy of language is directly relevant to understanding central issues surrounding artificial intelligence, our interaction with AIs, interpretable AI, and Explainable AI. Reflection about AIs and their representational capacities has proceeded in more or less complete isolation from the last 30 years of developments in philosophy of language and metasemantics. A tacit assumption is that our understanding of AI's representational and communicative abilities is best left to computer scientists and those who are trained to build the relevant kind of software.[1] The issues raised in this paper show that there's significant potential for philosophical engagement. We briefly illustrate potential payoffs in the final section with a discussion of existential risk, the value alignment problem, and explainable AI.

## 2. Alien Contents in AI

---

[1] At least, as far as we can tell. Looking at seminal survey papers in--to pick an area where one might expect some discussion of philosophy--explainable AI, one finds little to none contemporary philosophy either in overviews (Adadi & Berrada 2018, Mueller et al. 2019) or position papers (Doshi-Velez & Kim 2017), or even in work aiming to harness the social sciences to learn about explanation Miller (2019).

## 2.1 Types of AI Content

Some of the purposes to which we currently put, or hope soon to put, AI systems, involve communication with *our current contents*, but other applications will require contents that are different from the ones we have now. Let us illustrate both types before considering what to say about them.

*AIs with familiar contents:* When we build an image recognition/classification AI system, the goal is for its outputs to mean that the image is an image *of a duck* or *of a goose*. When we build a medical diagnostic AI system, the goal is that its outputs to mean that the tumor is *malignant* or *benign*. When we build a credit evaluation AI system, the goal is that its outputs mean that the client is *low-risk* or *high-risk*. For these sorts of systems to play the roles we want them to play, their contents need to be our contents, because we want them to answer specific questions we have, and that we have formulated using our contents. As we stressed in our (2021), there are difficult challenges in working out what about such AI systems would make their outputs have meanings, and have the specific meanings that we need them to have.

*AIs with unfamiliar contents:* Other applications of AI systems may require and give rise to contents that are not our contents. Consider an example. We find that average lifespans have been decreasing in a community, and we aren't able to work out why. So we feed enormous masses of data into an AI data-mining system: individual medical records, hospital records, weather patterns, economic indicators, crime reports and judicial rulings, social media activity, detailed traffic records from GPS tracking, and so on. The data mining system sifts the data performing linear regression and looking for significant patterns, and eventually identifies what it takes to be a crucial factor, pointing us to several vectors of that pattern (some hospital patients, a flock of migratory birds, a recent high-pressure system). It *could* be that the crucial factor is something nameable with our current contents (a known disease, a specific parasite thriving under certain environmental conditions, etc.). However, another possibility is that the crucial factor identified by the AI system *cannot be described in our language*. A relatively innocuous version of this would be a disease we had never detected and for which we thus had no name. Less innocuous versions might be causal factors that don't fall nicely into our existing categories of "disease", "violence", and so on, so that we would even find it hard to work out *what sort* of new thing we were looking for.

AI systems thus confront us with the possibility of *alien content*. This is not the first time theorists have considered that option. The process of radical translation/interpretation has always contained an element of alienation, although with the typical expectation that the contents, once uncovered, will prove to be familiar contents in alien garb. We return below to the connection between the tools of radical translation and the concerns about understanding alien AI contents. Kuhnian paradigm shifts also raise the spectre of alien contents -- the post-relativistic notion of *mass* is perhaps not wholly expressible in the pre-relativistic language. We glance toward the connections with Kuhnian incommensurability when we turn to issues of alien metasemantics below. In the data mining example, we at least know the logical category of the alien content. The AI system, by design, is expressing some categorizing predicate -- the difficulty is just that (ex hypothesi) our language contains no predicate, atomic or complex, matching the alien predicate in content. But at least the shared logical category gives us a place to start. Perhaps we can eventually learn the alien category by ostension of exemplars. Even if not -- if, for example, no collection of facts about extension of the alien category enable us to get onto its intension—there is a natural-enough similarity metric here, letting us perhaps find contents of our own that are similar enough to the alien content.

However, we may not always have the supporting crutch of a shared logical category. Consider the possibility of a map-making AI system, whose output is map-like representations of various places and things. Already we have taken away much of the familiar content-structuring infrastructure of our languages—the AI's maps need not use the tools of reference and predication, of quantification and boolean truth functions, of a broadly type-driven categorial semantics. And if the AI map-maker is given flexibility to redesign its own map-making methods, in quest of better representation of whatever is worth representing about the mapped environment, the similarity of its maps to our language-like conventions for creating map representations can dwindle away. The AI system shifts to projection methods using functions less and less natural to us, represents spatial relations in a non-Euclidean geometry or in a sub-Hausdorff topology, conflates spatial, temporal, and sociopolitical relations in its sub-topological organizational scheme, deploys map symbols holistically and inter-relationally, and so on. At some point, it would be unsurprising if the AI's maps were wholly incomprehensible to us, and untranslatable into anything we could recognize as a map or a meaningful representation.

The possibility of alien content is independent of the structure of the vehicle. While a predictive language system like GPT-3 produces output that wears the syntactic form of English, that syntactic form is no guarantee of semantic form. As we stressed in our (2021), the mere fact that an AI system outputs texts like "Lucie is high risk" does not mean, absent a controversial metasemantic argument, that the AI system's output *means that Lucie is high risk*. The AI could be using our words to express its contents. But, of course, the AI need not be using our syntax to express contents in a way guided by the syntactic structure at all. GPT-3 (if it means anything by its outputs) could be expressing contents without truth conditions, contents formally representable only as complex constraints on probability functions or other information measures, dynamic update rules on alien scoreboards, and so on. Again, there is no guarantee that anything "said" is something we could say or understand.

## 2.2 Communication (Or Communication*) With Alien Content Users

What should be done, then, were we to encounter an AI producing alien contents? Normally, when we encounter linguistic utterances we don't understand, we have reason to engage in a process of translation, looking for a method of mapping bits of their language to bits of our language in a meaning-preserving way. But of course translation is possible only when the same contents are available on the sending and the receiving ends, so that translation can match utterances with *the same content*. By hypothesis, that's not the situation we are in with alien contents.

A devotee of Quine and Davidson's method of radical translation might object that this hypothesis cannot be realized. Perhaps we can always run the radical translation machinery and always get an output, so we can always translate. There are two reasons why we find this an implausible idea, even from within the Quine-Davidson framework.

- First, despite the claims of Davidson (1973), even when the radical translation tools are ready to hand, they may produce no translation. We may recognize the alien activity as intelligent, guided by beliefs and desires -- activity classifying objects into categories, for example. We might nevertheless find that we have no contents matching the distinctions guiding the alien activity. We can find that no categories of ours categorize as they do.
- Second and more importantly, the gears of radical translation might fail to engage with alien contents. Depending on its flavor, radical translation requires in its object beliefs and desires, or rational action, or knowledge. AIs or other possessors of alien contents might have no such states. That's perhaps the expected situation for the kinds of AI systems we are familiar with: they have nothing like desires, and states only analogically like beliefs. They may be less possessors of knowledge and more tools for the expansion of *our* knowledge. One of the things that make AI systems in particular such an interesting case of alien content is that their natures are a mixture of the alien and the familiar. While they may not, for example, desire themselves, they are tools created by and serving satisfaction of our desires, and *something like* radical translation of them in light of their relations to us may be possible. We return to this theme below when we discuss meta-metasemantics.)

If translation is impossible because there is nothing of ours to translate their contents into, what is to be done? How is communication with or understanding of AI systems possible, given that communication and understanding also seem to require shared content? We'll explore two options, which we will call **bridging** and **integration**.

## 2.3. Bridging

Suppose an AI produces outputs with contents that we lack, so that we cannot simply translate its outputs into our language. We might nevertheless be able to build a bridge of understanding between us and the AI by finding suitable *corresponding* contents of ours.

Some versions of this strategy are familiar. Others have suggested that successful communication requires only that audiences uptake contents *similar* to contents entertained or intended by speakers, without requiring full identity.[2] Thus Frege finds it unlikely that two speakers will share the very same sense for a proper name, but thinks they can communicate so long as their senses are similar. (In the limiting case, perhaps shared denotation suffices for similarity.)

---

[2] See, for example, Carston 2001, Sperber and Wilson 1986, Bezuidenhout 2002).

Or bridges could be built using the mereology of content.[3] A verbal description of the contents of a painting, for example, might share content with the painting by capturing some of what is represented in the painting, but unavoidably also leave out some of the pictorial content because pictorial representation makes available other contents that simply can't be produced with linguistic tools. Perhaps, similarly, the alien content of the AI is essentially richer than our content, adding some who-knows-what on top of our familiar contents. But if we can, when encountering the AI output, at least get onto the portion of its content that overlaps with ours, some form of communication will have been achieved. Or the mereology might run in the other direction. Perhaps we have the richer contents, and the AI system can only produce contents that are coarsenings or diminishing of ours. Again, a form of communication is available by associating a class, or a representative of a class, of our richer contents with the AI's alien diminished content.

But the really interesting cases are ones in which we aren't so fortunate as to have a ready-made bridging relation like similarity or parthood. How do we *build* a bridge to alien contents, when it's a new bridge that's needed? Nothing of full generality can be expected here, but we can point to what we think are the crucial questions and tools.

It's helpful to get started here to realize that the aliens are already among us, and that they are in fact us. Philosophers are adept at producing creative and radically varying pictures of the nature of content, so philosophers disagree wildly among themselves about what sort of contents they are producing. These philosophers can't all be *right* (since they disagree), but we can still consider how communication would happen between communities properly modelled by different philosophical pictures of content. Consider two test cases:

*First, consider the expressivist*:[4] Simplifying, the expressivists take utterances not to represent the world, but rather to express, to give literal voice to, various mental states. Perhaps we are all expressivist about "ouch" -- its sole content is as a tool for the expression of pain. More thoroughgoing expressivists may view our moral language as expressive of conative or evaluative attitudes, or our modal vocabulary as expressive of global features of our doxastic state, or the vast swath of our descriptive vocabulary as expressive of our beliefs. The contents of the expressivists cannot be *translated* into the contents of the representationalist -- the representationalist traffics in propositions, while the expressivist does not, so there is no common ground between them. In some sense, each of the expressivist and the representationalist should be licensed in simply ignoring the other, if taken on the other's own terms, since by their lights the other is saying nothing they can comprehend.

But there is a natural enough *mapping* of expressivist to representationalist content. We can see that already with "ouch". We say "ouch" to express our pain, but instead of expressing our pain we can report on it, by saying "I am in pain". Anything that is expressed can similarly be reported on. ("I have a con attitude toward murder", etc.) The reportings don't say the same thing as the expressions, but the relation between reporting and expressing can support a lot of our communicative practice. The representationalists who takes the expressivists' expressions as reportings, or the expressivist who takes the representationalists' reportings as expressings, won't get everything right, but will have an interpretational practice that will allow them to make considerable sense of the other.[5]

---

[3] See e.g. Yablo 2014, Fine 2016 and the literature that it has spawned.
[4] Gibbard 1990, Schroeder 2008 and references therein.

*Second, consider the Lewisian centered-worlds de se theorist:*[6] The fan of centered worlds takes contents to be properties, not propositions. When he says "I am hungry", his utterance has the property of *being hungry* as content; when he says, "Brutus stabbed Caesar", his utterance has the property of *being B-related and C-related to a pair of objects, the first of which stabbed the other* as content. Again, the fan of possible worlds content cannot translate the utterances of the centered-worlds speaker, since his contents are all propositions, and not properties. In some sense fans of centered worlds and fans of uncentered worlds should regard each other's view proclamations as simply incomprehensible, by virtue of not having (by their own lights) the right sort of content.

As in the expressivist case, there is a natural enough mapping between centered and uncentered content. Centered content does not have truth conditions (being properties rather than propositions), but there is a natural way for the fan of propositions to extract associated truth conditions. On Lewis's view, what we do, cognitively, when we have an attitude whose content is a property is to *self-ascribe* that property. That self-ascription has accuracy conditions -- one rightly or wrongly self-ascribes a property. So the uncentered theorist can map the centered expression of property P by an agent A to the uncentered content *that A is P* via the correctness conditions of A's self-ascribing P. The centered theorist can similarly extract a centered content from every uncentered content *that phi*, by mapping it to the property of *occupying a phi world*.

Perhaps these mappings don't allow for communication, in the strict sense, between expressivist and representationalist, or between centered and uncentered theorist, but they allow for some form of communication*—something that is communication-like enough to be recognized. How do we find the right mapping to allow for communication*, especially as the contents to be mapped from get increasingly alien? One way to answer this question is to look to the theoretical role that content plays. One formal gadget or another (sets of worlds, sets of centered worlds, etc) counts as a content because it does something for us. Because our fellow philosophers are much like us, we can look for familiar roles for (e.g.) expressivist or centered contents -- explaining action, guiding assessments of rationality, allowing modification of the beliefs of others, and so on. A representationalist has one view on the sort of things best positioned to play those roles, so when encountering an expressivist content, they can look for a content of their own that's suited to play the role in the way the expressivist's content, on the expressivist worldview, is.

---

[5] These sorts of bridging by content-mapping strategies are related to, but importantly distinct from, the paraphrase strategies that get used in ideal language metametaphysics. When the mereological nihilist paraphrases the ordinary language "there is a chair" as "there are some simples arranged chairwise", he finds a suitable analog for the non-nihilist content among the contents he accepts, much as we are suggesting that the representationalist finds a suitable analog for the expressivist content among her own contents, or (below) that we find suitable analogs to the alien AI contents among our own contents. However, the mereologist nihilist and non-nihilist share contents -- the nihilist has the content "there is a chair", but thinks that that content is false, so wants to map onto another content (one also available to the non-nihilist) that she thinks is true. And the paraphrase has the stable goal of finding a paraphrase that matches the assertability conditions of the target sentence, whereas the kind of bridging maps we are considering can have highly variable success conditions. (Note that alien contents need not have assertability conditions that we can express any more than they have truth conditions that we can express, and may not have assertability conditions at all, because they may not be used in a practice of assertion.)

[6] Lewis 1979.

The sketch we just gave of the theoretical role of content is deeply anthropocentric. The challenge faced in extending this approach to alien contents of AI systems is to generalize and de-anthropocentrize our understanding of the theoretical roles that contents play. Only with a de-anthropocentized account can we get a clear picture of what kinds of contents can play the relevant role for AIs. This line of thought thus leads to an investigation of metasemantics -- what features make things mean what they do -- and the metasemanics suitable for AI systems. We turn to that investigation in section 3. Before doing that, we consider the second option to encountering alien contents: integration.

## 2.4 Integration

Rather than seeking a suitable content of our own to match with the AI's alien contents, we could *acquire* the alien contents, integrating ourselves into their linguistic practice and making ourselves users of those contents. Consideration of simple cases might suggest that the process of integration is straightforward. When our diagnostic data-mining AI gets onto a new category, sorting objects in some way previously unavailable in our language, we can simply introduce a new predicate and stipulate that its intension matches the intension by which the AI is sorting. The integration move here is a deferential move, but it's important to distinguish two forms of deference:

- In *semantic deference*, the property of deference is part of the meaning of the deferential expression. "Gloobish", in our language, *means* "has the feature that the AI system picks out when its red light flashes".
- In *metasemantic deference*, on the other hand, the property of deference is no part of the meaning of the deferential expression -- rather, deference is a metasemantic feature that fixes the nondeferential meaning. If the AI system is in fact picking out property P, then with metasemantic deference, our "gloobish" simply means that P, by way of our deference.

Semantic deference is easily achieved (as easily as any other stipulation). However, semantic deference is also too cheap for most communicative purposes. With semantic deference, we can stipulate that "iliadic" means *whatever the Greek expression* "μῆνιν ἄειδε θεὰ Πηληϊάδεω Ἀχιλῆος//Οὐλομένην" *means*. If we have no idea what it does mean, "iliadic" gives us only thin arms-length access to the content, not genuinely integrating it into our language. It gives us nothing we didn't already have via quotation.

Metasemantic deference is what we want. That's the tool that allows the alien contents to *become* contents of our language. Metasemantic deference, however, is much less obviously easily achieved. We need to know what kind of deferential relation is needed to make our words mean what the AI words mean. There are at least two worries here. First, *our* notion of deference may not apply to AIs. Perhaps to defer is in part to respect the target of deference, to think that the target knows more about the subject than we do. But AI systems may not be sensible objects of respect and may not have knowledge as we understand it. If so, perhaps we *can't* defer to them. And second, our notion of deference may not be the right one. Perhaps human deference is a metasemantic ground of meaning in human languages—but why think that human deference grounds meaning acquisition between human and alien languages. Perhaps alien deference is needed, or perhaps the alien language is entirely deference-proof.

Of course, deference isn't essential here—it was only one possible route to integration. But the concern generalizes. What we want is (to use the Lewisian (1969) terminology) to put ourselves in the actual language relation to the alien contents. But to do that, we need to know what it takes to stand in the actual language relation to that sort of content. That is, we need to know about the metasemantics of the alien language. If the alien contents are grounded (externalist-style) in causal interactions with other speakers of the alien language, perhaps integration is not too hard -- we simply interact with the AI systems in the right way for a time. If the alien contents are grounded (conventionalist-style) in participating in conventions of trust and truthfulness, integration may be hard to impossible, since AI systems may not be capable of entering into the relevant sorts of conventions.

## 3. Alien Metasemantics

There is something that makes our words and utterances mean what they do. There is much philosophical dispute about what that something is perhaps—intentions of speakers, perhaps conventions within a community of language users, perhaps teleofunctional facts, perhaps causal connections to the environment, and so on. Whatever these metasemantic[7] facts are that give our words meaning, their role is metaphysical, not epistemological. For the most part, *figuring out* what someone's utterance means is something we do without consulting the metasemantics -- we understand someone's utterance of "Aristotle was fond of dogs" without engaging in historical investigation of the chain of usages lying behind their utterance. However, when we encounter difficulties in interpretation, turning to the metasemantics to work out content can be one of the tools open to us.[8]

One of the major themes of our (2021) is that we ought to attend to metasemantic questions when considering questions about the content and intelligibility of AI systems. In that book, we weren't primarily concerned with metasemantics as a tool for figuring out what was meant.

---

[7] We're terminologically following Robert Stalnaker here who says that metasemantic questions are, "about what the facts are that give expressions their semantic values, or more generally, about what makes it the case that the language spoken by a particular individual or community has a particular descriptive semantics" (Stalnaker 1997: 535). Stalnaker's paradigm of a metasemantic theory is Krikpe's causal theory of proper names.

[8] Of course, it's contentious what the right metasemantic view is, so difficulties in interpretation may simply spill over into difficulties in selecting a metasemantic framework. And even when we agree on the metasemantics, the method of metasemantic determination may not be scrutable by us.

Rather, the emphasis was on attending to the metasemantics in order to convince ourselves that, or evaluate whether, our AI systems are indeed producing meaningful outcomes. In addition, considering the metasemantics of AIs can help us determine whether they are similar to us in their content production in a way that would allow us to use our standard epistemic methods of determining content among ourselves (methods that, again, don't typically run through the metasemantic determination) to the AIs as well.

Once we are confronted with the possibility of alien contents in AI systems, the epistemological stakes are raised, and we will need all the tools we can find to help achieve understanding—or, following the discussion of the previous section, whatever analog of understanding is available. We suggested in that discussion that attention to the metasemantics of AIs could be useful either to help us figure out what mapping relation between alien and domestic contents provided the best replacement for translation or to help determine how we needed to position ourselves relative to AIs to integrate their content production with ours. So what might such attention to the metasemantics reveal?

## 3.1 Metasemantic Contingency

It's helpful before discussing the metasemantics of AI to remind ourselves that metasemantic facts are contingent facts, which depend on details of the language and the language community for their truth. Disputes between internalists and externalists are often framed as if one of these two positions is a necessary truth capturing some essential fact about the nature of content and communication. But surely this isn't right. Classic arguments for metasemantic externalism—in, for example, Kripke (1980) and Burge (1979)—start with particular contingent facts about our languages. We speak a language whose content is fixed externally in part *because* we judge that name-involving utterances depend on the person at the origin of the causal chain of name uses, rather than on the person satisfying the cluster of descriptions entertained by the speaker, or because we are disposed to defer to experts when they correct our referential assumptions. If we behaved differently—disregarded the causal history, declined to defer—the externalist arguments would fail, and an internalist metasemantics would fit our language. Surely we *could* have behaved differently, so surely the metasemantics could have been otherwise.[9]

## 3.2 Varieties of Metasemantic Variability

Once we acknowledge that there can be variability in the metasemantics, we can identify possible dimensions of that variability. We might distinguish:

---

[9] There are complicated questions here about whether our various possible linguistic behaviors would be *constitutive* of an internalist or externalist metasemantics, or merely *evidence for* an internalist or externalist metasemantics. To think the first would plausibly be to assume a background internalism about the metasemantics—an internalist metametasemantics. We return to these higher-order questions below.

- *Lexical variation in metasemantics*: perhaps names have their contents fixed through one mechanism, sentential connectives through another, moral vocabulary through a third.
- *Syntactic variation in metasemantics*: perhaps names as used in atomic sentences rely on one metasemantics, while names in attitude contexts rely on another.
- *Linguistic variation in metasemantics*: perhaps meanings in English are fixed in one way, while meanings in C++ are fixed in another way.
- *Speaker variation in metasemantics*: perhaps demonstratives as used by some speakers have their referents fixed by speaker intentions, while demonstratives as used by other speakers have their referents fixed by discourse coherence relations.

These various dimensions of metasemantic variability are independent—in principle, any combination of them could be manifested. Given our interest in looking at the metasemantics of AIs in order to gain traction on problems of alien content, we will focus primarily on interspeaker metasemantic variation, but the other categories of metasemantic variation deserve further investigation as well.[10]

## 3.3 Why AI-metasemantics will be (to some extent) alien

As we stressed in our (2021), it's not hard to see that an adequate metasemantics for AI content producers would likely need to be different from an adequate metasemantics for content producers like us. The details depend on one's preferred metasemantics, of course, but to rehearse a few examples:
- If our metasemantics has it that meanings of utterances are (à la Grice (1989)) determined by the communicative intentions of speakers, but AI systems lack intentions of any sort, then we need a different metasemantic story for those AI systems (perhaps appealing to the deferred communicative intentions of the AI programmers?)
- If our metasemantics has it that meanings of utterances are (Davidson-style) determined by what best rationally mediates between perceptual inputs and action outputs for speakers, but AI systems lack perceptual faculties or capacities for action, then we need a different metasemantic story for those AI systems (perhaps replacing perception with data input and action with data output?).

Our strategy in our (2021) was to find appropriate anthropocentric abstractions of existing metasemantic theories, so that both AI and human content production could be seen as instances of a more generalized metasemantic theory. Thus we emphasized the potentially important role of externalist metasemantic theories in thinking about AI content attribution, while at the same time stressing that *existing* exteralist frameworks had their specific externalisms shaped by thinking exclusively about content generation for humans.[11] If, however, the AI systems become different enough from us, there may be no possibility of finding a single metasemantic story explaining the production of content in both us and the AI systems. This is a familiar lesson from consideration of multiple realizability. There may be no single grounding story that explains both why mad pain is pain and why Martian pain is pain (Lewis 1980).

---

[10] Maybe we should never talk about variation in metasemantics, but instead look for a generalized or parameterized metasemantics that spans across putative areas of variation? So if we're initially driven to think that names have their contents determined in one way and connectives have their contents determined in another way, we should instead take the data to suggest that there is a single metasemantic framework that tells us that (e.g.) *being a name* combined with *having a particular causal history* grounds one content while *being a connective* combined with *having a particular inferential role* grounds another content. It's not clear how much beyond terminological boundary drawing is at issue here. In some sense we could look for a single grounding story for absolutely everything, not just contents, that maps each "lower-level" collection of facts to its corresponding grounded "higher-level" fact. The question is just whether theoretically interesting generalizations about grounding within certain areas thereby get lost or obscured. Talk of multiple realizability can, in this context, be interpreted as recognition of important differences in grounding generalizations within the same general sort of stuff.

So we may need to confront, in addition to the possibility of alien contents, the possibility of alien metasemantics. Consider an example. When the predictive language system GPT-3 produces textual output, it produces output that has (let us suppose) some content. What then determines the content of GPT-3 outputs? Given that GPT-3 outputs have the same syntactic and morphological structure as strings in our English, one option is to transfer the English-language metasemantics to GPT-3 outputs (assuming that is possible)—perhaps standing linguistic conventions fix the meanings of both our sentences and GPT-3's sentences. Note, however, that this transfer move isn't mandatory. We could instead consider a fundamentally alien metasemantics for GPT-3. For example, GPT-3's neural net weightings are determined by extensive training against massive text databanks provided by the Common Crawl archiving service. Give the role that the Common Crawl database plays in the training of GPT-3, maybe the Common Crawl archiving algorithms play a metasemantic role for GPT-3 as well, so that what GPT-3 means by "snow is white" is determined (perhaps in some way we can't fully comprehend) by the way in which Common Crawl extracts and stores text content from various websites.[12]

## 3.5 When is an Alien Metasemantics too Alien?

If alien metasemantics gets *too* alien, we might wonder whether it is meta*semantics* at all. Suppose Alex says that their pet rock right now is saying that it's snowing. Beth suggests that this just can't possibly be right. There's nothing about the pet rock that would make it *say* something. It doesn't have any beliefs or intentions. It's not a participant in any conventions, so a conventionalist metasemantics won't make it be saying anything. It's not in a causal chain of uses, so a Kripkean externalist metasemantics won't make it be saying anything. It's not properly interpretable as saying that it's snowing, so an interpretationalist metasemantics won't make it be saying anything. And so on—none of our metasemantic stories make it say anything at all.

---

[11] AI systems have the potential to challenge not only the way we think about external factors playing a role in content determination, but also the way we draw the line between internalist and externalist metasemantic framework. A conception of "internal duplicate" that is (perhaps) well-understood in the human case may become much more elusive when we extend it to the AI case. Are two copies of the same neural net implemented on different computers with (for example) different computational power internal duplicates? If an AI system regularly accesses other computational tools (database lookup tables, etc), do those other tools count as "internal" or "external" to the AI? Questions of this sort are already complicated for us by the extended mind hypothesis, but become particularly obscure when extended to AIs.

[12] Note that alien contents and alien metasemantics are orthogonal issues: An AI could have an alien metasemantics, but via those alien metasemantic mechanisms end up producing contents that are our familiar contents. Similarly, an AI could have alien contents produced by a familiar metasemantics. And, of course, an AI could have both -- alien contents generated by an alien metasemantics.

Alex says:

> Well, of course not. You're trying to use human meta-semantic stories on a rock. That's just not the right way to go about it. Rocks come to have content in very different ways than the ways people come to have contents. Look, do you see how there's this cluster of quartz crystals there on the upper left of the rock? That's what makes it the case that it's saying that it's snowing. That's because the metasemantics for rocks is that having quartz crystals in various places is what makes it the case that it's saying things.

Alex's response is unconvincing. While we should be open to alien metasemantic determination, surely not just any old collection of properties can give rise to content. So the proponent of an alien metasemantics has an explanatory debt: they need to make sure that their alien metasemantic story really is plausible as a *metasemantic* story. A mere variant on the pet rock story won't discharge that explanatory debt.

How can the debt be paid, then? One approach starts by looking at the kinds of things contents are, and then figuring out what it would take to ground those things. An analogy: take some economic event-type, like a recession. We can ask for a grounding story for a recession (think of this on analogy with "metasemantics of a recession"). We want to know in virtue of what a recession is happening. And there could be lots of stories here—high unemployment, stagnant wage growth, high interest rates, and so on. Furthermore, we could consider the possibility of an "alien metasemantics of recession"—we could imagine creatures who didn't have banks or jobs, and for whom recessions weren't matters of high interest rates or high unemployment rates. But we couldn't just appeal to any old feature of these creatures—it's not going to make sense that for them, a recession is a situation in which they all need haircuts. What we need to do is think about what kind of thing a recession is—some kind of slow-down in economic growth. Then we can look at these creatures and ask what makes for economic growth for them. They don't have banks, because they don't have currency. Instead, it's all barter for them—but we could then imagine that changes in the bartering practices amount to a recession for them. We can see that the bartering changes work and the haircut changes don't work by thinking about the nature of the thing we're trying to ground.

Alien metasemantics can be approached in the same vein. Suppose that content is just truth conditions. Then whatever our metasemantics is, it needs to be the kind of thing that can make sense of giving rise to truth conditions. Human metasemantic stories do explain that—maybe they involve tracking differences in reacting to different kinds of situations, and extracting truth conditions from the differences in the situation kinds. But the distribution of quartz crystals in the rock just isn't the kind of thing that's going to be able to explain differences in truth conditions, so it can't be a good metasemantic story for a truth-conditional account of content.

That response to the "pet rock" challenge depends on a shared notion of content. With mild versions of alien contents, there could still be enough sharing to get this response going. If the AI is alien only insofar as it has predicates that can't be translated into open sentences of ours, but familiar insofar as it makes truth-conditional claims by applying (alien) predicates to familiar objects, we might be able to leverage the shared truth-conditionality to recognize the AI's alien metasemantic machinery as properly metasemantic. But when the alien contents become more thoroughly alien, this response will fail.

When the contents are alien, then, we can't look to the nature of content to convince us that a putative alien metasemantic story is plausibly meta*semantic*. Where do we look then? Our suggestion: just as uncertainty about alien contents can be tackled by looking to the metasemantics, uncertainty about the alien metasemantics can be tackled by looking to the *meta-metasemantics*.

## 3.6 Meta-Metasemantic Guidance to Metasemantics

Consider a hierarchy of explanation. We start with facts about content. The name "Aristotle" has the semantic feature of picking out a certain individual. We then ask: *why* does "Aristotle" have that feature (rather than some other semantic value)? The metasemantics answers this question: "Aristotle" refers to Aristotle because Aristotle lies at the origin of the causal chain of usages terminating in the current use of "Aristotle". Then we ask again: *why* does the causal chain determine the content of "Aristotle", rather than some other feature (such as the descriptions associated by the speaker with the name "Aristotle")? The meta-metasemantics answers this question.[13] The meta-metasemantic options are less thoroughly mapped than the metasemantic options. Here is an example of how to think about a meta-metasemantic option:

> *Informational-Epistemic Role Meta-metasemantics*: It's the causal chain that determines the meaning rather than the associated descriptions *because we use names as tools for transmitting information and accumulating knowledge about distant objects whose properties we track only imperfectly. A causal chain metasemantics gives contents to those names that let them play those informational and epistemic roles better than a descriptive metasemantics would*.

There are many other meta-metasemantic options, but for now we'll use this as an illustration of how it can guide us in understanding when something is a metasemantic mechanism.

## 3.7 An Illustration of Meta-metasemantic Guidance

---

[13] Metametasemantics can be connected to talk of grounding in two ways: (i) We can take the meta-metasemantics to ground the grounding claim provided by the metasemantics. (ii) We can take the metametasemantics to be further elaborating the grounding claim given by the metasemantics, by giving further conditions left implicit in the original grounding claim or by specifying *enabling conditions* (e.g. Dancy 2004) that allow the grounds to ground. So, roughly, within a grounding context the meta-metasemantic facts can be accommodated by moving either *upward* (to the grounding of the grounding) or *outward* (to a broader perspective on what is doing the grounding). We remain neutral here on the best way to connect our meta-metasemantic talk to the grounding literature.)

Suppose we encounter a sophisticated AI system that has been designed to help guide the global economy. Unfortunately, it is very obscure to us. We are told the following:

> *Potential Alien Content + Alien Metasemantics*: (i) The Economic AI outputs have as their contents alien contents that are *probability distributions on sets of strategies for playing certain games*. (ii) These alien contents are plausible because we can see how they would be grounded in simulation games that the AI is playing both internally and with a collection of other game-playing AI systems, combined with facts about the way in which certain major stock market indices typically react to outputs of the AI.

Suppose we agree that these facts about simulation games plausibly ground a relation between the Economic AI and the probability distributions. This raises the question of why we should think of these probability distributions on sets of strategies as *contents*—that is, why we should think of the grounding story as a *metasemantic* grounding story. In answering this, we appeal to the kind of meta-metasemantic considerations outlined above:

> *Meta-metasemantic guidance:* The *point* of contents is to allow for rational assessment of our action and of our internal transition from one content-bearing state to another. The Economic AI doesn't perform *human actions* or engage in *human-style belief updating*, but based on its design we know that it is to be assessed based on how well it guides our economy and on how well it internally configures itself to spot opportunities for guiding our economy. The (alien) contents attributed on the basis of the Economic AI's internal and external game-playing and of the stock index reactions, the suggestion goes, make best sense of the AI as assessable on those grounds, so that attribution basis gives the right metasemantics for the AI.

The point here is that looking to the point of contents—the role they play in our larger theorizing about language, action, cognition, and so on—gives us a sketch of a meta-metasemantics. The metasemantics can ground *contents*, rather than something else, because being so grounded makes sense of being able to play the right theoretical roles. Some plausible anthropocentric abstraction then allows us to take a human-centered meta-metasemantics and see how to adapt it to produce a meta-metasemantics, not thoroughly alien, that covers both us and the AI system.

## 3.8 Meta-metasemantics and Shared Ways of Life

If a lion could talk, Wittgenstein tells us, we could not understand him (*Philosophical Investigations*, p223). Its form of life, we take Wittgenstein to be suggesting, is too different from ours for there to be common ground for communication. We don't know quite what a form of life is, but we can offer our own version of this thought:

> *From Ways of Life to Meta-metasemantics:* If the metametasemantics is going to guide access to alien metasemantics and alien contents, then the theoretical roles for contents in the AI need to be similar enough to the theoretical roles for contents for us. But the theoretical roles for contents for us depend on a bunch of contingencies about the sorts of creatures we are and the sorts of things we do. That we have perceptual faculties, that we have interests in tracking specific objects over time and space, that we trade information among ourselves, that our actions are explained by the quality of our evidential relations to the world, and so on—all of these things fix the theoretical role of contents, and all of these things could have been different.

We're actually reasonably optimistic about hypothetical lion talk—they seem a lot like us in many of the relevant ways. But there's a range of cases here. Even more like us are our pets—they share much of our physiology, but also some of our sociology as well, unlike lions. Much less like us is, for example, the oceanic life form of Stanislaw Lem's *Solaris*, which doesn't obviously have anything at all similar to our goals or world-tracking mental states. The utility of looking to the meta-metasemantics will vary with the similarity of the alien content-bearer to us, and different dimensions of similarity and difference can each bring their own challenges.

The intriguing thing about AI systems, then, is that they have a unique combination of similarities to, and differences from, us. In some ways, even the most sophisticated AI is much less like us than is a lion. The AI (plausibly) doesn't have beliefs and desires, appetites and fears, or direct control over a physical body. But in other ways, the AI is much more like us than even our pets are. It, like us, contains neural nets.[14] The AI has epistemic resources that are closely analogous to what we have: they are trained on, e.g. visual inputs, they discriminate various properties from each other, they use these 'perceptual' capacities to classify situations into those that, say, have stop signs in them and those that don't . These discriminatory capacities are familiar to us.[15] After all, the systems were made to do things that can be useful to us: for example, distinguish ducks from not-ducks, malignant tumors from benign, and to classify people according to their credit worthiness. So at the core of the systems, we find familiar abilities, based on mechanisms (neural nets) that we share with them. In short: we say that these systems are fundamentally different, but not alien.

Hopefully, then, these similarities are enough to enable us to find theoretical roles for AI content. More specifically, hopefully we can use a familiar metametasemantics to make sense of an alien metasemantics, and then make use of that alien metasemantics to figure out how to map alien contents to our own contents or how to integrate ourselves into the alien contents of the AI. The de-anthropocentrizing may be difficult. Perhaps knowledge acquisition features in the theoretical role of content, but perhaps AI systems aren't the kind of thing that can *know*. In our (2021) we suggested one way of performing the anthropocentric abstraction here: hold on to the role of knowledge, but shift from the language *producers* as knowing to the language *consumers* as knowing. Another option is to generalize our epistemic taxonomy, to find some more general evidence-tracking state of which knowledge is one instance (the instance found in us) and some other state is another instance (the one found in the AIs).

---

[14] It also includes features (e.g. the use of neurotransmitters and hormones) that artificial neural networks lack. So we don't describe a brain anywhere near exhaustively by calling it a neural net. A helpful discussion here is Garson 2008: 4, but the key point is that there's at least an important structural similarity but also an important dissimilarity between human brains and neural nets and that suffices to make our point.

[15] Plausibly the point carries over to , for example, credit-assessing systems, or systems concerned with recidivism. Although they needn't take sensory inputs, it nevertheless processes the same sort of information about people we do when we assessing people.

## 3.9 When there's no common meta-metasemantics: meta-meta-metasemantics to the rescue!

Hope may eventually give out. Maybe the theoretical role of content in the end depends on categories that have no analog in AI systems. Suppose contents are those things that explain action and rational accessibility as evidence managers. Certain kinds of AIs have nothing analogous to action or to evidence management. What then? Of course, eventually we might become convinced that the entire ecosystem of content-based talk is inapplicable to AIs that differ enough from us. We're not there, however, just because we can't find a common meta-meta-semantic level . We can continue to ascend the meta-ladder. When appeal to a shared meta-metasemantics gives out, we can turn to the meta-meta-metasematics. We won't examine this option in any detail here—it's hard enough to think about what makes it the case that certain features make it the case that a speaker means particular things by what they say, without also trying to think about what makes it the case that those things make it the case that certain features make it the case that a speaker means particular things by what they say. We can perhaps get started by casting the "theoretical role" net more broadly—perhaps, just as we looked to the theoretical role of content in explaining action and rational assessment in order to get a grip on meta-metasemantics, we can look to the theoretical roles of action and rational assessment in order to get a grip on meta-meta-metasemantics.

There is no in-principle limit to how far we might have to ascend the meta-ladder in order to find a point of adequate commonality to begin a process of communicative bridging to sufficiently odd AIs. This can be a dizzying thought. It's natural to think that there must be a fixed point somewhere—that what we want is communication, which requires shared content, and that if we can't have that, there's nothing to have; or that what we want is knowledge, which requires something communication-like (but not necessarily communication in the sense of shared content), and that if we can't have that, there's nothing to have. What would it be like to engage with beings for whom there was no fixed point at which the question of how to engage was set?

# 4. Concluding Applications: Explainable AI and Existential Risk

The issues we've discussed here can easily seem abstract and of no relevance to the real and pressing practical, social, and political issues raised by the issues that AI presents us with. Nothing could be further from the truth. These are foundational issues with very direct practical implications. We provide two illustrations: Existential Risk and Explainable AI.

## 4.1. Existential Risk

There has been a great deal of literature on "existential risk" concerns about AI (see Bostrom 2014 for a popular recent discussion). These concerns often center on what is called The Alignment Problem: how to get the goals of AI systems sufficiently aligned with ours, and how we might design AIs to guarantee a comforting level of alignment. We have preferred to focus on a collection of what we take to be underexplored semantic and metasemantic questions about AI systems, but the two topics intersect when we start thinking about alien content.

Here is a simple version of the existential risk issue: we might want some version of Asimov's First Law of Robotics (first presented in his 1942 story "Runaround") incorporated into any AI systems we produce:

> Asimov's First Law, AI version: No (AI system) may injure a human being, or by inaction allow a human being to come to harm

Implementing exactly that law in an AI system requires shared content—if the AI's semantic resources don't include the concepts of *injury* or *harm*, then we won't be able to tell it what to do. If an AI has alien contents, then we can't share that content with the AI (because Asimov's First Law is formulated using human contents.) One immediate corollary is that settling the issue of alien vs. non-alien content is a requirement on even thinking about the alignment issue.

Another corollary of the above discussion is that the presence of alien contents doesn't mean giving up on the First Law of Robotics. Suppose we've settled that the AI can't share our concept of *injury or risk*. How do we recognize the alien concepts as being suitably related to *harm* and *injury*? One option suggested above is that we can look to the metasemantics or the meta-metasemantics. We check to see if the alien concept is metasemantically determined by facts related to human flourishing; we consider what theoretical role the distinction between flourishing and non-flourishing lives plays for us. Pursued correctly, such discussions bring together questions of ethics, content, and meta-content into a single research project.

## 4.2 Explainable AI

The aim of XAI is to create AI systems that are *interpretable* by us, that produce decisions that come with comprehensible *explanations*, that use *concepts* that we can understand, and that we can *talk to* in the way that we can engage with each other. The importance of this is highlighted by recent legislation that makes it a requirement that decisions made by (or with the help of) neural nets come with an *explanation* of the decision. Thus--and without getting too deep into the weeds of contemporary EU data protection law[16]--the EU's enshrines a 'right to explanation'. One has the right to "meaningful information about the logic [of the algorithm] involved."[17] That means that those subject to neural net-based decision-making are owed an explanation of how a decision affecting them was arrived at. This is obviously a very tall order when we don't even know whether an AI has the same kind of contents as we humans have. Most of the discussion of explainable AI proceeds on the assumption that we are dealing with a system that generates outputs with human-type contents. Even then the requirement that an explanation be available is hard to meet, but the challenge gets even harder once we recognize the real possibility that the relevant outputs could contain alien contents generated by alien metasemantic mechanisms. We then need procedures for recognizing these contents as contents, and we need ways to relate to those alien contents. This paper has provided proposals for how to proceed, but it should be obvious that we have at best scratched the surface of some extremely hard questions and that there's much work to be done.

---

[16] See Goodman and Flaxman 2017 for discussion
[17] Articles 13 and 14 of the GDPR, quoted in Goodman and Flaxman 2017:6.